# Inverse-Category-Frequency based Supervised Term Weighting Schemes for Text Categorization

DEQING WANG, AND HUI ZHANG
*State Key Laboratory of Software Development Environment*
*Beihang University*
*Beijing, 100191 P.R.China*
*{dqwang, hzhang}@nlsde.buaa.edu.cn*

Term weighting schemes often dominate the performance of many classifiers, such as kNN, centroid-based classifier and SVMs. The widely used term weighting scheme in text categorization, i.e., *tf.idf*, is originated from information retrieval (IR) field. The intuition behind *idf* for text categorization seems less reasonable than IR. In this paper, we introduce inverse category frequency (*icf*) into term weighting scheme and propose two novel approaches, i.e., *tf.icf* and *icf-based* supervised term weighting schemes. The *tf.icf* adopts *icf* to substitute *idf* factor and favors terms occurring in fewer categories, rather than fewer documents. And the *icf-based* approach combines *icf* and relevance frequency (*rf*) to weight terms in a supervised way. Our cross-classifier and cross-corpus experiments have shown that our proposed approaches are superior or comparable to six supervised term weighting schemes and three traditional schemes in terms of macro-F1 and micro-F1.

## 1. INTRODUCTION

Text Categorization (TC—a.k.a. text classification), is the task of labeling natural language texts with thematic categories from a predefined set [21]. A natural language text is often converted into a machine friendly format so that the computer or classifiers can "understand" the content of the text. This step is called text representation. In the vector space model (VSM), the content of a text is represented as a vector in the term space, i.e., $d = \{w_1,...,w_K\}$, where $K$ is the term (feature) set size [9]. The term weight $w_i$ indicates the degree of importance of term $t_i$ in document $d$. The term weighting schemes often affect the effectiveness of classifiers. For example, Leopold and Kindermann [10] pointed out that the performance of SVM classifiers is dominated by term weighting schemes, rather than kernel functions. Therefore, a well-defined term weighting scheme should assign an appropriate weighting value to each term.

At present, there are two types of term weighting schemes: unsupervised term weighting schemes (UTWS) and supervised term weighting schemes (STWS). The UTWS are widely used for TC task [4, 10, 13, 21]. The UTWS and their variations are borrowed from information retrieval (IR) field, and the most famous one is *tf.idf* (term frequency and inverse document frequency) proposed by Jones [7, 8]. Robertson [19]

Received; revised ; accepted.
Communicated by Deqing Wang.
* This work was supported supported by the 863 High-Tech Program under Grant No. 2007AA010403

tried to present the theoretical justifications of both *idf* and *tf.idf* in IR.

However, the TC task differs from the IR task. For TC task, the categorical information of terms in training documents is available in advance. The categorical information is of importance for TC task. Debole and Sebastiani [3] proposed supervised term weighting methods that used the known categorical information in training corpus. They adopted the values of three feature selections (i.e., $\chi^2$, information gain, and gain ratio) to substitute *idf* factor during weighting terms. Their thorough experiments did not exhibit a uniform superiority with respect to standard *tf.idf* [3]. Supervised term weighting schemes, however, seem more reasonable than unsupervised ones for TC task.

Recently, Liu *et al.* [13] proposed a probability based supervised term weighting scheme (denoted as *prob-based*) to solve imbalanced text classification problem. Similarly, Lan *et al.* [9] used relevance frequency (*rf*) to substitute *idf* and also proposed a novel supervised term weighting, i.e., *tf.rf*. However, both supervised term weighting schemes have a common shortcoming, which is the simplification of a multiclass classification problem into multiple independent binary classification problems. During the process of the simplification, the distribution of a term among categories disappears because there are only positive category and negative category.

Previous researches have shown that *tf* is very important and using *tf* alone can achieve good performance for TC task [9, 10, 21]. Therefore, researchers focus on *idf* factor instead. The discriminating power of a term in document *d* not only is related to *tf*, but also is related to the distribution of the term among categories. The intuition is: ***the fewer a term appears in categories, the more discriminative the term is for text categorization***. It is similar to *idf* in IR task. In this paper, we inspect the role of inverse category frequency (*icf*) and introduce *icf* into term weighting scheme for TC task, then we propose two novel term weighting schemes based on *icf*, i.e., *tf.icf* and *icf-based* supervised term weighting schemes, which are applied to multi-class, and binary TC, respectively. On three widely used TC corpora, i.e., the skewed Reuters-21578, the balanced 20 Newsgroup, and la12, our *tf.icf* outperforms three traditional schemes (*tf*, *idf*, and *tf.idf*) in terms of macro-averaging and micro-averaging F1 with three different classifiers on multi-class classification tasks. Moreover, our *icf-based*, adopting the merits of *rf* and *icf*, outperforms seven supervised term weighting schemes (e.g., *tf.rf* and *prob-based*) and standard *tf.idf* on binary classification tasks.

The rest of the paper is organized as follows. We briefly review related work in Section 2. Then we present an overview of term weighting schemes, including unsupervised term weighting schemes and supervised term weighting schemes in Section 3. In Section 4, we propose *tf.icf* and *icf-based* supervised term weighting scheme separately, and give detailed explanation. The text collection, text process, benchmark methodology of classifiers, and evaluation measures are presented in Section 5. We discuss detailed experimental results on three corpora in Section 6, and we conclude this paper in Section 7.

## 2. RELATED WORK

In recent years, TC has been widely used in many applications, for instance, spam email classification, question categorization and online news classification. Generally, TC tasks can be divided into single-label learning task and multi-label learning task [21].

Many TC techniques have been explored in the literature, e.g., centroid-based classifier [4-5], kNN [12, 23], Naïve Bayes [14], decision tree [18] and support vector machines [2, 6]. Since this paper focuses on term weighting functions, those who are interested in TC techniques can find details from Sebastiani [21].

Term weighting functions originate in IR field, one of which is the famous *tf.idf* with huge success in IR. Salton and Buckley [20] discussed many typical term weighting schemes in IR field, and they found *tfc* (normalized *tf.idf*) is the best document weighting function. Because of *tf.idf*'s success in IR, researchers retain the function for TC task, and many TC tasks usually take *tf.idf* as the defaulted term weighting function. There are some new improved schemes. For instance, Xue and Zhou [22] found distributional features (the compactness of the appearance of a term) were useful for text categorization. Guan *et al.* [4] applied inter-class and inner-class term indices to construct centroid vectors with better initial values than traditional methods. And the inter-class index is a variation of *icf* factor.

However, Debole and Sebastiani [3] proposed the supervised term weighting schemes idea by replacing *idf* factor with three feature selection functions, i.e., information gain, chi-square, and gain ratio. Lan *et al.* [9] proposed *tf.rf* to improve the performance of text categorization, which is only related to relevant documents. Their experiments showed *tf.rf* outperformed other supervised term weighting schemes (such as *tf.logOR*, $tf.\chi^2$, *tf.ig*) and traditional ones (such as *tf.idf, tf, binary*). Moreover, Liu *et al.* [13] proposed a probability based supervised term weighting scheme to improve the performance of imbalanced text classification. Quan *et al.* [17] adopted three new supervised term weighting schemes (i.e., *qf*icf, iqf*qf*icf* and *vrf*) for question classification. Pei *et al.* [16] proposed an improved *tf.idf* method, which combined *tf.idf* and information gain. Many variations of ICF have been used in document classification [26, 27, 30, 31].

Despite of the vast research efforts on term weighting schemes, further study is still needed to reveal whether inverse category frequency can be beneficial to term weighting schemes and how to realize it. Our study in this paper just tries to make some contributions to this problem.

## 3. OVERVIEW OF TERM WEIGHTING SCHEMES

### 3.1 Unsupervised term weighting schemes

As mentioned above, the unsupervised term weighting schemes are borrowed from IR field. The intuition of *idf* in IR filed is that a query term with occurrence in many documents is not a good discriminator, and should be given less weight than those that have occurred in less documents[19]. Debole and Sebastiani [3] concluded three assumptions about *tf.idf* model, i.e., (i) *idf assumption*: rare terms are no less important than frequent terms; (ii) *tf assumption*: multiple occurrences of a term in a document are no less important than single occurrence; (iii) *normalization assumption*: long documents are no more important than short documents for the same quantity of term matching. According to these assumptions, researchers proposed many variations of *tf.idf* model. Here we present its standard "ltc" variation [20].

$$tfidf(t_i, d_j) = tf(t_i, d_j) \times \log \frac{|Tr|}{\#Tr(t_i)} \tag{1}$$

where $|Tr|$, $\#Tr(t_i)$ denote the total number of documents, the number of documents containing term $t_i$ in training set $Tr$, respectively. $tf(t_i, d_j)$ has various variations such as raw tem frequency, *binary* or *log(1+tf)*. In our study, we use raw term frequency as $tf(t_i, d_j)$. To eliminate the effect of document length, the $L_2$-normalization is performed to normalize the term weight obtained by Eq. (1).

$$w_{ij} = \frac{tfidf(t_i, d_j)}{\sqrt{\sum_{k=1}^{|V|} \left[ tfidf(t_k, d_j) \right]^2}} \tag{2}$$

where *|V|* denotes the total number of unique terms contained in training set *Tr*.

Many researchers believe that term weighting schemes in the form as *tf.idf* can also be applied into TC task. The intuition of *idf* in IR filed seems reasonable. However, it needs to be carefully weighed for TC task because TC task is different from IR task. For TC task, a term, which occurs in many documents, maybe is a good discriminator, especially when it occurs in a few categories or only one category. For example, in the corpus of patents, each patent in category "computer science" may contain the term "computer" or "software". If the document frequency of the term in corpus is very high, does it mean that the term has less discrimination? In contrast, the term can be regards as a powerful feature about category "computer science". Therefore, ***we should consider the distribution of a term among categories, rather than among documents***. This is our first motivation to revise the term weighting scheme for TC task.

### 3.2 Supervised term weighting schemes

Researchers have noted that *tf.idf* may not be the best term weighting model for TC task. Therefore, Debole and Sebastiani [3] proposed supervised term weighting schemes. They introduced term (feature) selection into term weighting schemes and made the phases of term weighting to be an activity of *supervised learning*, in which information on membership of training documents in categories is used. Because later research showed that the models using feature selection metrics (such as $\chi^2$, information gain, gain ratio, Odds Ratio, and so on) to weight term is not superior to traditional *tf.idf* model [9, 13, 33], we do not give detailed formulations, readers who are interested in the methods can refer to Debole and Sebastiani [3].

**Table 1. Fundamental information elements used for supervised term weighting schemes.**

|           | $c_j$ | $\overline{c_j}$ |
|-----------|-------|------------------|
| $t_i$     | $a$   | $c$              |
| $\overline{t_i}$ | $b$   | $d$              |

Recently, two new supervised schemes are proposed to deal with binary classification, i.e., *tf.rf* and *prob-based*. The fundamental information elements used for term weighting scheme are shown in Table 1, where $a$ is the number of documents in $c_j$ (positive category) that contain term $t_i$; $b$ is the number of documents in $c_j$ that do not contain term $t_i$; $c$ is the number of documents in $\overline{c_j}$ (negative category) that contain term $t_i$; $d$ is the number of documents in $\overline{c_j}$ that do not contain term $t_i$. The *tf.rf* [9] is expressed as

$$tf.rf = tf \times \log(2 + \frac{a}{\max(1,c)}) \qquad (3)$$

In addition, the *prob-based* scheme [13] is as follows

$$prob-based = \frac{tf(t_i, d_j)}{\max(tf(d_j))} \times \log(1 + \frac{a}{c} \times \frac{a}{b}) \qquad (4)$$

where $\max(tf(d_j))$ is the maximum frequency of a term in document $d_j$.

The two schemes have a common factor, i.e., *a/c*. It means the weight of term $t_i$ is related to relevant documents that contain this term $t_i$. There are four different factors between Eq. (3) and Eq. (4), including (i) The base of logarithmic operation of *rf* factor in Eq. (3) is 2, whereas the base is $e$ (natural number) in Eq. (4); (ii) the constants in Eq.(3) and Eq. (4), are 2, and 1, respectively; (iii) *tf* factor in Eq. (4) is divided by $\max(tf(d_j))$, whereas *tf* in Eq. (3) is raw term frequency; (iv) Eq. (4) considers the ratio between relevant documents and irrelevant documents in specific category $c_j$, i.e., *a/b*.

As we know, *prob-based* scheme is proposed to deal with imbalanced text classification, so *a/b* is supposed to balance the rare categories. For instance, given two categories $c_i$ (major category, which contains 100 documents) and $c_j$ (minor category, which only contains 3 documents.), if $a_{c_i} = 50$, $b_{c_i} = 50$ and $a_{c_j} = 2$, $b_{c_j} = 1$, then $a_{c_i}/b_{c_i} < a_{c_j}/b_{c_j}$. The term weighting value in $c_j$ is improved by the factor *a/b*. This will improve the probability of assigning a document to $c_j$. However, introducing the parameter *b*, i.e., the number of documents which do not contain term $t_i$ in a specified category $c_j$, has no intuitive explanation. As the authors presented, *prob-based* scheme gave more weight to terms occurred in minor category. In our experiments on skewed Reuters-21578, *prob-based* scheme achieved worse performance. However, after removing the factor *a/b*, the *macro-F1* and *micro-F1* on imbalanced Reuters-21578 corpus improve obviously. Through our experiments, we can verify that the *a/b* factor is less important than the *a/c* factor and it may affect the performance of a classifier.

The *tf.rf* scheme outperforms *prob-based* because it only considers the frequency of relevant documents. However, *tf.rf* favors terms that frequently occur in the positive category. It is opposite to *idf*, which favors rare terms. For instance, if a term $t_i$ only occurs once in category $c_j$, $t_i$ should be a good discriminator for category $c_j$. However, the $rf(t_i) = \log(2 + 1/\max(1,0)) = \log(3)$ Following Zipf's law [24] of regular words, the frequency of many terms in a big corpus should be one, and *rf* assigns less weight to these terms. This will weaken the discriminating power of these words.

Besides, *tf.rf* and *prob-based* schemes both have one shortness, i.e., ***the distribution of a term among categories disappears during partitioning training corpus into positive and negative categories***. As we know, a term occurring in fewer categories should have more discriminating power, but *tf.rf* and *prob-based* schemes do not consider the factor. This is our second motivation to revise the term weighting scheme.

## 4. TWO NOVEL TERM WEIGHTING SCHEMES BASED ON ICF

In this section, we introduce *inverse category frequency* (*icf*) into term weighting schemes for TC task. Two concepts are defined as

- ◆ *Category frequency* (*cf*): the number of categories in which term $t_i$ occurs.
- ◆ *Inverse category frequency* (*icf*): the formula of *icf* is similar to *idf* and it is expressed as

$$icf(t_i) = \log(\frac{|C|}{cf(t_i)}) \tag{5}$$

where $|C|$ denotes the total number of categories in the training corpus.

The *icf* has been widely used in many TC tasks. For instance, Quan *et al.* [17] used a variations of *icf* as one factor in their question categorization problem. Guan *et al.* [4] adopted *icf* to construct centroid vector of category for their centroid-based classifier that outperformed SVM classifiers on closed tests. The intuition behind *icf* is: ***the fewer categories a term occurs in, the more discriminating power the term contributes to text categorization***. This assumption is named ***icf assumption*** in which i*cf* favors rare terms and biases against popular terms on the category level.

Therefore, we introduce *icf* factor into term weighting scheme and propose two novel approaches, i.e., *tf.icf* and *icf-based* supervised term weighting schemes. The formulae are expressed as Eq. (6) and Eq. (7), respectively.

$$tf.icf(t_i, d_j) = tf(t_i, d_j) \times \log(1 + \frac{|C|}{cf(t_i)}) \tag{6}$$

$$icf-based(t_i, d_j) = tf(t_i, d_j) \times \log(2 + \frac{a}{\max(1, c)} \times \frac{|C|}{cf(t_i)}) \tag{7}$$

With respect to *tf.idf*, the *tf.icf* is a mixture model of term weighting. Because the two factors in *tf.idf* model are both estimated on the document level, but the *tf* factor of *tf.icf* is estimated on the document level and the *icf* factor is estimated on the category level. Our *tf.icf* is different from previous *tf.icf* methods [25-31], which have the same name, but have different meanings. For example, Reed [25] use the abbreviation *ICF* (Inverse Corpus Frequency) in dealing with stream data, and ICF in Ref [25] is inverse document frequency of the whole corpus. Lertnatteed [26, 27] also proposed TFICF, whose *TF* factor is estimated on category level. Kimura *et al.* [30] proposed TF.ICF to weight terms in cross-language retrieval system, and the TF factor [30] is defined as the ratio between term frequency and the number of terms in category $c_j$. Lei *et al.* [31]

adopted DF factor and ICF factor to weight terms in Web document classification. The similar methods can be found in Refs [28-29].

The *icf-based* scheme contains three factors. The *tf* factor is the raw term frequency; *rf* factor measures the distribution of term $t_i$ between positive category and negative category; *icf* factor measures the distribution of term $t_i$ among categories. The *cf* information can be acquired and saved when analyzing training corpus, thus it can be used to calculate term weight after training corpus is partitioned into positive category and negative category.

For further clarification of two supervised term weighting schemes (*icf-based* and *tf.rf*), consider the terms $t_1$, $t_2$ having the distributions given in Table 2, and suppose $t_1$ occurs in fewer categories than $t_2$, i.e., $cf(t_1) = 2$ and $cf(t_2) = 4$. According to Eq. (3), we know $tf.rf(t_1) = f.rf(t_2)$, because *tf.rf* only considers the distribution of a term in positive and negative categories. However, the discriminating power of $t_1$ should be larger than that of $t_2$, because $t_1$ occurs in fewer categories than $t_2$. According to Eq. (7), we can find $icf-based(t_1) > icf-based(t_2)$. It means *icf-based* supervised term weighting scheme seems more reasonable than *tf.rf*. Our experiments on real-world datasets will verify our assumption again.

**Table 2. The document/category frequency of term $t_1$ and $t_2$.**

|  | $t_1$ | $t_2$ |
|---|---|---|
| *df* in positive class | 10 | 10 |
| *df* in negative class | 5 | 5 |
| *cf* in the entire training set | 2 | 4 |

**Table 3. Case study of nine term weighting schemes on category *acq*.**

|  | *icf-based* | *rf* | *prob-based* | *OR* | $\chi^2$ | *gr* | *ig* | *icf* | *idf* |
|---|---|---|---|---|---|---|---|---|---|
| acquir (9) | 5.543 | 3.335 | 1.568 | 30.524 | 1191.792 | 0.711 | 0.167 | 2.306 | 0.887 |
| stake (9) | 5.301 | 3.206 | 1.011 | 23.161 | 680.111 | 0.427 | 0.100 | 2.788 | 1.098 |
| payout (1) | 1.0 | 1.0 | 0 | 2.797 | 89.462 | 0.100 | 0.023 | 3.246 | 2.397 |
| dividend (2) | 1.540 | 1.024 | 4.207 | 0.080 | 184.795 | 0.223 | 0.052 | 2.324 | 1.791 |

**Table 4. Case study of nine term weighting schemes on category *earn*.**

|  | *icf-based* | *rf* | *prob-based* | *OR* | $\chi^2$ | *gr* | *ig* | *icf* | *idf* |
|---|---|---|---|---|---|---|---|---|---|
| acquir (9) | 1.357 | 1.068 | 0.001 | 0.083 | 416.694 | 0.276 | 0.095 | 2.306 | 0.887 |
| stake (9) | 1.402 | 1.079 | 0.001 | 0.104 | 231.742 | 0.163 | 0.056 | 2.788 | 1.098 |
| payout (1) | 13.508 | 7.820 | 3.004 | 652.736 | 239.146 | 0.138 | 0.047 | 3.246 | 2.397 |
| dividend (2) | 9.544 | 4.936 | 2.052 | 36.101 | 557.825 | 0.325 | 0.112 | 2.324 | 1.791 |

As Lan *et al.* did in Ref [9], we select four terms (i.e., "acquir", "stake", "payout", and "dividend") from real-world text corpus to verify the effectiveness of *icf-based* method. The first two terms are related to category **acq** and the last two terms are related to category **earn**. Table 3 and 4 list the values of terms using nine term weighting schemes. The number nearby each term is the category frequency in the corpus. As table 3 and 4 shown, the weighting values of four terms are the same when using *idf* and *icf*, because the two methods do not consider positive and negative categories. The term "payout" is filtered when *earn* is tagged as the positive category and *ig*, *gr*, or $\chi^2$ is used. *Prob-based* method assigns more weighting value to the term "dividend" wrongly when *acq* is tagged as positive category. The remaining methods, i.e., *OR*, *.rf*, and our *icf-based* can discriminate terms between the two categories correctly. It is worthy noting that the category frequency of "payout" is one, that is, the term only appears in category **earn.** The weighting value of "payout" is enlarged to 13.508, which emphasize the role of *cf*.

## 5. EXPERIEMTNAL SETUP

### 5.1 Text corpora

**Reuters-21578:** The Reuters-21578 data set[*] is widely used benchmarking collection [3-6, 9, 11, 23, 33]. Our data set is based on the Trinity College Dublin version, which changed documents from the original SGML format into XML format. According to the "*ModApte*" split, we got a subset of 52 categories after removing unlabeled documents and documents with more than one class labels, denoted by *Reuters-52*, which is a single-label multi-class corpus [4]. There are 6532 training documents and 2568 test documents. The imbalance problem of *Reuters-52* is more serious, the most common category (*earn*) accounts for 43% of the whole training set, whereas the bottom 10 categories of it only contain several training instances in each category. 319 stop words, punctuation and numbers are removed; all letters have been converted into lowercase; word stemming is not applied. The final vocabulary has 27,953 words.

**20 Newsgroup:** The data set[†] consists of 19,905 documents, uniformly distributing in twenty categories. The data set is also a famous benchmark in TC tasks [4, 5, 9]. We randomly select 33% instances from each category as test instances and the rest texts as training instances. There are 13,330 training instances and 6,575 test instances. We only keep "Subject", "Keywords", and "Content". Other information, such as "Path", "From", "Message-ID", "Sender", "Organization", "References", "Date", "Lines", and email addresses, are filtered out. The stop words list [15] has 823 words, and we keep words that occur at least twice. All letters are converted into lowercase and word stemming is not applied. The final vocabulary has 24,162 words.

**La12：** The corpus[‡] is the larger data set in text-data collection. It is derived from TREC and consists of 6279 examples. We randomly choose 25% examples as test set and the rest as training set. The total number of words is 31,472.

---

For all data sets, no feature selection is used because the three classifiers can handle high-dimensional data. For Reuters-21578, we run one trial because it provides separated training and test sets according to "ModApte" split; For 20 Newsgroup and La12, we do 10 trials since there is no given test set, and we acquire the averaged macro-F1 and micro-F1of 10 trials to evaluate our proposed approaches.

## 5.2 Text Process

Before applying classifiers, text is represented as a vector. The text process between UTWS and STWS is different, thus we give the detailed explanations as follows. For UTWS, we compare their performance on multi-class classification task. That is, we first process the training set, and save the *idf* (or *icf*) value of each term into files, which will be used to calculate the weighting value of each term in test set. Then the $L_2$-normalization (Eq. 2) is performed to normalize the term weight.

For STWS, we evaluate the nine methods on binary classification task, because they are suited to binary classification [3,9]. That is, in each experiment, a chosen category $c_j$ is tagged as the positive category, and the rest categories in training corpus are combined together as the negative category. Then we obtain the statistical information (*a, b, c, d* and *cf*) of each term in positive and negative classes, and save them into files. When analyzing the document from test set, we calculate the weighting value of each term by combining its *tf* and the statistical information in training set.

## 5.3 Classifiers

We choose the state-of-art algorithm, i.e., SVM-based classifier, because SVM almost achieves top-notch performance among the widely used classification algorithms [6, 9, 21, 22]. We use LIBSVM [1] and adopt two kernel functions, i.e., linear function and radial basis function (RBF). For RBF and multi-class classification tasks, we use 5-fold cross validation to find the optimal parameters $C$ and $\gamma$, where $C$ is the penalty parameter and $\gamma$ is the kernel parameter of RBF. The cross validation is time-consuming, thus for binary classification tasks, we select linear kernel and defaulted parameters in order to save training time.

Meanwhile, centroid-based classifier outperforms kNN, Naïve Bayes and C4.5 for text categorization according to previous research, and centroid-based classifier will be affected by term weighting scheme [5]. kNN is also adopted in order to compare with previous research [9, 13], and we set k=10 for each experiment, as Han et al. did in Ref [5]. The similarity measure we use for the classifiers is the cosine function. Unless otherwise specified, we use the default parameter values for each classifier in our experiments.

## 5.4 Performance Measures

We measure the effectiveness in terms of precision (*p*) and recall (*r*) defined in the usual way [11]. The two measures are popular performance measures for TC tasks. As a measure of effectiveness that combines the contributions of *p* and *r*, we use the well-known F1 function [11], defined as

$$F1 = \frac{2p \times r}{p + r} \tag{8}$$

Usually, *F1* is estimated from two ways for multi-class problem, i.e., macro-averaging F1 (macro-F1) and micro-averaging F1 (micro-F1) [3, 4, 9]. The macro-F1 gives the same weight to all categories, thus is mainly influenced by the F1 of rare categories for skewed Reuters-25718 corpus. On the contrary, micro-F1 will be dominated by the performance of common categories for skewed Reuters-25718 corpus. Therefore, the macro-F1 and micro-F1 of Reuters-25718 may give quite different results. Because of the balance of 20 Newsgroup, the macro-F1 and micro-F1 of 20 Newsgroup are quite similar. For binary classification problem, we only obtain the true-positive number to calculate F1 score of positive class, and then use them to calculate the macro-F1 and micro-F1 of the entire data set.

## 6. RESULTS ANALYSIS

We have constructed a number of groups of experiments to verify the performance of *tf.icf* and *icf-based* supervised term weighting schemes.

### 6.1 Comparisons of UTWS on multi-class classification task

Firstly, we compare *tf.icf* with unsupervised term weighting schemes on **multi-class classification** tasks. We report the overall performance of *tf.icf* method on three corpora for multi-class classification task. Tables 5 and 6 show the performance of four term weighting schemes in terms of macro-F1 and micro-F1 on skewed *Reuters-52* with four classifiers, and the best result is highlighted in bold.

**Table 5. The macro-F1 comparison of four term weighting schemes on skewed Reuters-52.**

|       | SVM(RBF) | SVM(LINEAR) | kNN   | Centroid |
|-------|----------|-------------|-------|----------|
| tf    | 0.699    | 0.585       | 0.527 | 0.631    |
| idf   | 0.581    | 0.507       | 0.522 | 0.662    |
| tfidf | 0.736    | 0.654       | 0.634 | 0.679    |
| tficf | **0.739** | **0.655**  | **0.671** | **0.712** |

As shown in Table 5 and 6, the *tf.icf* performs consistently the best in all experiments, especially for the kNN and centroid-based classifiers. For example, compared with *tf.idf*, the macro-F1 and micro-F1 of *tf.icf* using centroid-based classifier improve up to 3.3%, 1.4%, respectively; and using kNN, the improvements are 3.7%, 4.8%, respectively. When using SVM classifier, the performance of *tf.idf* is very close to *tf.icf*. The reason is that the performance of SVM is dominated by kernel function, rather than term weighting function. As shown in the second and third columns, SVM with RBF is sig-nificantly better than SVM with linear kernel.

**Table 6. The micro-F1 comparison of four term weighting schemes on skewed Reuters-52.**

|       | SVM(RBF) | SVM(LINEAR) | kNN   | Centroid |
|-------|----------|-------------|-------|----------|
| tf    | 0.935    | 0.924       | 0.856 | 0.821    |
| idf   | 0.909    | 0.904       | 0.816 | 0.866    |
| tfidf | 0.934    | 0.930       | 0.818 | 0.855    |
| tficf | **0.937**| **0.931**   | **0.866** | **0.869** |

The same results can be observed from Table 7, which reports the performance on balanced 20 Newsgroup corpus. Because the macro-F1 and micro-F1 are very close, we only report the micro-F1 in Table 7. Compared with traditional term weighting methods, we find that *tf.icf* achieves the best performance. With respect to *tf.idf*, the micro-F1 of *tf.icf* has improved up to 2% when using the centroid-based classifier; and the improvement is 2.5% when using kNN.

Table 8 and 9 show the performance comparisons of four term weighting schemes on la12 data set. We can observe that our *tf.icf* outperforms other three methods in terms of macro-F1 and micro-F1. The improvement is significant when kNN and centroid-based classifiers are used. For example, comparing to *tf.idf*, the macro-F1 and micro-F1 of *tf.icf* are improved up to about 5% when employing kNN.

**Table 7. The micro-F1 comparison of four term weighting schemes on 20 Newsgroup.**

|       | SVM(RBF) | SVM(LINEAR) | kNN   | Centroid |
|-------|----------|-------------|-------|----------|
| Tf    | 0.833    | 0.827       | 0.774 | 0.742    |
| idf   | 0.837    | 0.826       | 0.807 | 0.758    |
| tfidf | 0.839    | 0.830       | 0.788 | 0.796    |
| tficf | **0.843**| **0.837**   | **0.813** | **0.806** |

**Table 8. The macro-F1 comparison of four term weighting schemes on la12.**

|       | SVM(RBF) | SVM(LINEAR) | kNN   | Centroid |
|-------|----------|-------------|-------|----------|
| tf    | 0.913    | 0.899       | 0.768 | 0.793    |
| idf   | 0.902    | 0.890       | 0.797 | 0.852    |
| tfidf | 0.919    | 0.907       | 0.817 | 0.846    |
| tficf | **0.926**| **0.913**   | **0.864** | **0.864** |

**Table 9. The micro-F1 comparison of four term weighting schemes on la12.**

|  | SVM(RBF) | SVM(LINEAR) | kNN | Centroid |
|---|---|---|---|---|
| tf | 0.930 | 0.922 | 0.800 | 0.817 |
| idf | 0.922 | 0.913 | 0.813 | 0.876 |
| tfidf | 0.930 | 0.932 | 0.845 | 0.870 |
| tficf | **0.933** | **0.935** | **0.896** | **0.889** |

The *tf* and *idf* achieve worse performance in all four classifiers, which has been verified in the previous research [9, 13]. We also find the performance of *tf* is superior to that of *idf* when SVM and kNN classifiers are used, and the conclusion is opposite when the centroid-based classifier is adopted.

We employ the McNemar's significance test [32] to verify the difference on the performance of two term weighting schemes. On Reuters-52 corpus, (*tf.icf*, *tf.idf*) $\gg$ *tf* $\gg$ *idf* when using SVM classifier, where "$\gg$" denotes better than at significance level 0.01. However, *tf.icf* $\gg$ *tf* $\gg$ (*tf.idf*, *idf*) when using kNN classifier, and (*tf.icf*, *idf*) $\gg$ *tf.idf* $\gg$ *tf* when employing centroid-based classifier.

On 20 Newsgroup corpus, (*tf.icf*, *tf.idf*) $\gg$ (*tf*, *idf*) when SVM classifier is used, and *tf.icf* $\gg$ *tf.idf* $\gg$ *idf* $\gg$ *tf* when employing kNN, and (*tf.icf*, *tf.idf*) $\gg$ *idf* $\gg$ *tf* when using centroid-based classifier.

On la12 data set, (*tf.icf*, *tf.idf*) $\gg$ (*tf*, *idf*) when SVM classifier is employed, and (*tf.icf*, *idf*) $\gg$ *tf.idf* $\gg$ *tf* when employing kNN, and *tf.icf* $\gg$ (*idf*, *tf.idf*) $\gg$ *tf* when using centroid-based classifier.

According to our cross-classifier and cross-corpus experiments, *tf.icf* should be used as the standard term weighting scheme for multi-class TC task, because *tf.icf* seems more reasonable than *tf.idf* and *tf.icf* can fully exploit the known information of training instances, i.e., the distribution of keyword among categories. Meanwhile, *tf.icf* can achieve consistently best performance in our all experiments.

## 6.2 Comparisons of STWS on binary classification task

In this section, we continue to show the superiority of *icf-based* supervised term weighting scheme for **binary classification** task, compared with 8 existing methods, i.e., *tf.rf* [9], *prob-based* [13], *tf.logOR* [9], $tf.\chi^2$ [3], *tf.gr* [3], *tf.ig* [3], *tf.icf*, and *tf.idf*. For supervised term weighting schemes, we adopt "local policy" [9] to construct training and test sets, because the supervised term weighting methods are suited to binary classification tasks. That is, in each experiment, a chosen category $c_j$ is tagged as the positive category, and the rest categories in training corpus are combined together as the negative category.

Table 10 and 11 report the overall macro-F1 and micro-F1 of nine term weighting schemes on skewed Reuters-52 corpus, respectively. As shown in Table 10, our proposed *icf-based* method consistently outperforms eight methods on three classifiers in terms of macro-F1, and the improvement of *icf-based* method is significant. For example, when using SVM, compared with $tf.\chi^2$, *tf.gr*, the performance of *icf-based* method improves up 15%, 11% , respectively; And we also observe that *icf-based* is always superior to *prob-based* and *tf.idf* in terms of macro-F1. Then we compare the performance between *icf-based* method and state-of-the-art *tf.rf*. The best macro-F1 of *icf-based* supervised term weighting scheme on *Reuters-52* corpus reaches up to 54.7%. The *icf-based* is a little superior to *tf.rf* , and *icf-based* method improves 0.8% than *tf.rf*. Meanwhile, we observe that our *tf.icf* achieved the third best macro-F1 score when using SVM, and *tf.ig*, *tf.gr*, *tf·$\chi^2$*, and *prob-based* methods achieved worse performance.

When using centroid classifier, our *icf-based* method achieves the best macro-F1, i.e., 35.2%, which is improved up to 10% than *tf.rf* method. The reason is that *tf.rf* performs lower F1 on minor categories. Then *prob-based* and *tf·$\chi^2$* methods achieve the second best macro-F1 score. The rest methods are worse than *icf-based* method.

When using kNN, we can obtain the similar conclusion. Our *icf-based* method is a little better than *tf.rf* with 1% improvement on macro-F1. The conclusion using kNN is consistent with Lan's in Ref [9].

**Table 10. The macro-F1 comparison of nine term weighting schemes on skewed Reuters-52 corpus.**

|  | icf-based | tf.rf | prob-based | tf.logOR | $tf·\chi^2$ | tf.gr | tf.ig | tf.icf | tf.idf |
|---|---|---|---|---|---|---|---|---|---|
| SVM | **0.547** | 0.539 | 0.395 | 0.455 | 0.391 | 0.435 | 0.435 | 0.494 | 0.473 |
| kNN | **0.504** | 0.494 | 0.486 | 0.455 | 0.428 | 0.470 | 0.470 | 0.394 | 0.350 |
| Centroid | **0.352** | 0.252 | 0.337 | 0.231 | 0.320 | 0.257 | 0.257 | 0.265 | 0.240 |

**Table 11. The micro-F1 comparison of nine term weighting schemes on skewed Reuters-52 corpus.**

|  | icf-based | tf.rf | prob-based | tf.logOR | $tf·\chi^2$ | tf.gr | tf.ig | tf.icf | tf.idf |
|---|---|---|---|---|---|---|---|---|---|
| SVM | **0.919** | 0.900 | 0.808 | 0.876 | 0.813 | 0.862 | 0.862 | 0.890 | 0.883 |
| kNN | **0.822** | 0.816 | 0.787 | 0.754 | 0.805 | 0.781 | 0.781 | 0.721 | 0.666 |
| Centroid | 0.947 | 0.954 | 0.938 | 0.949 | 0.952 | 0.946 | 0.946 | **0.960** | 0.958 |

In terms of micro-F1, the similar results are shown in Table 11, and *icf-based* supervised term weighting scheme performs better results in 2 out of 3 classifiers, except centroid classifier, which achieves the best micro-F1 with our another method, i.e., *tf.icf*. For example, when using SVM, *icf-based* method performs the best accuracy, i.e., 91.9%, which is 1.9% bigger than *tf.rf* method. When centroid classifier is used, *tf.icf* method achieves the best performance.

Table 12 shows the overall micro-F1 performance of nine term weighting schemes on balanced 20 Newsgroup corpus, which is a little different with that on *Reuters-52*. As shown in Table 12, *icf-based* method, *tf.icf*, and *tf.logOR* achieve the best micro-F1 when using SVM, kNN, and centroid-based classifiers, respectively. When SVM is used, the best *micro-F1* of *icf-based* reaches 73.6%, which is improved about 2% than *tf.rf*.

**Table 12. The micro-F1 comparison of nine term weighting schemes on 20 Newsgroup.**

|          | icf-based | tf.rf | prob-based | tf.logOR | $tf \cdot \chi^2$ | tf.gr | tf.ig | tf.icf | tf.idf |
|----------|-----------|-------|------------|----------|-------------------|-------|-------|--------|--------|
| SVM      | **0.736** | 0.716 | 0.562      | 0.728    | 0.529             | 0.526 | 0.526 | 0.718  | 0.727  |
| kNN      | 0.636     | 0.704 | 0.527      | 0.701    | 0.464             | 0.505 | 0.505 | **0.765** | 0.722 |
| Centroid | 0.833     | 0.829 | 0.794      | **0.931** | 0.822            | 0.791 | 0.791 | 0.882  | 0.909  |

It must be noted that *tf.icf* with kNN and *tf.logOR* with Centroid classifier significantly outperform our *icf-based* method with SVM classifier. The reason could be that the skewed data affects the performance of SVM classifier. Because the ratio between positive category and negative category is 1:19, and we use defaulted parameters to train SVM classifier for binary classification tasks. SVM is a global optimization scheme which may lead to the misclassifications of instances of minority classes [34, 35], On the contrary, kNN and centroid classifiers have some advantages on dealing with imbalanced text classification [5,13,21]. We also noted that the property of data corpus has a great impact on term weighting schemes. For instance, *tf.idf* achieves the second best performance on 20 Newsgroup, even better than *icf-based* scheme when kNN or centroid-based classifier is used. The similar explanation can be found in Ref [9].

The macro-F1 and micro-F1 on la12 corpus are shown in table 13 and 14, respectively. We can observe that our *icf-based* method outperforms other methods in terms of macro-F1 and micro-F1. For example, compared with *tf.rf*, *icf-based* method improves up to 3.2% when using kNN on micro-F1; and the improvement is up to 3.9% when employing centroid-based classifier. Besides *tf.rf*, the performance of *icf-based* method is significantly better than that of other methods, e.g., *tf.idf*, *tf.logOR*, and *tf.ig* etc.

**Table 13. The macro-F1 comparison of nine term weighting schemes on la12.**

|          | icf-based | tf.rf | prob-based | tf.logOR | $tf \cdot \chi^2$ | tf.gr | tf.ig | tf.icf | tf.idf |
|----------|-----------|-------|------------|----------|-------------------|-------|-------|--------|--------|
| SVM      | **0.872** | 0.866 | 0.668      | 0.818    | 0.760             | 0.739 | 0.739 | 0.866  | 0.864  |
| kNN      | **0.828** | 0.800 | 0.566      | 0.696    | 0.697             | 0.668 | 0.668 | 0.803  | 0.776  |
| Centroid | **0.810** | 0.778 | 0.563      | 0.697    | 0.609             | 0.605 | 0.605 | 0.772  | 0.775  |

**Table 14. The micro-F1 comparison of nine term weighting schemes on la12.**

|          | icf-based | tf.rf | prob-based | tf.logOR | $tf \cdot \chi^2$ | tf.gr | tf.ig | tf.icf | tf.idf |
|----------|-----------|-------|------------|----------|-------------------|-------|-------|--------|--------|
| SVM      | **0.884** | 0.872 | 0.656      | 0.831    | 0.762             | 0.752 | 0.752 | 0.871  | 0.866  |
| kNN      | **0.846** | 0.814 | 0.590      | 0.725    | 0.677             | 0.668 | 0.668 | 0.801  | 0.779  |
| Centroid | **0.899** | 0.860 | 0.661      | 0.793    | 0.798             | 0.767 | 0.767 | 0.869  | 0.860  |

Based on the McNemar's significance test [32], on Reuters-52 corpus, we observe that *icf-based* >> *tf.rf* >> (*tf.icf*, *tf.idf*, *tf.logOR*) >> (*tf.ig*, *tf.gr*) >> ( *tf.$\chi^2$*, *prob-based*) when using SVM classifier at significant level 0.01. When using kNN, (*icf-based*,

*tf.rf*) $>>$ *tf.* $\chi^2$ $>>$ (*prob-based*, *tf.gr*, *tf.ig*) $>>$ *tf.logOR* $>>$ *tf.icf* $>>$ *tf.idf*. When employing centroid-based classifier, (*tf.icf*, *tf.idf*, *tf.rf*, *tf.* $\chi^2$) $>>$ (*icf-based*, *tf.logOR*, *tf.gr*, *tf.ig*) $>>$ *prob-based*.

On 20 Newsgroup data set, *icf-based* $>>$ (*tf.idf*, *tf.logOR*) $>>$ (*tf.icf*, *tf.rf*) $>>$ *prob-based* $>>$ *tf.* $\chi^2$ $>>$ (*tf.ig*, *tf.gr*) when adopting SVM; and when using kNN, *tf.icf* $>>$ *tf.idf* $>>$ (*tf.logOR*, *tf.rf*) $>>$ *icf-based* $>>$ *prob-based* $>>$ (*tf.gr*, *tf.ig*) $>>$ *tf.* $\chi^2$. When centroid-based classifier is employed, *tf.logOR* $>>$ *tf.idf* $>>$ *tf.icf* $>>$ (*icf-based*, *tf.rf*, *tf.* $\chi^2$) $>>$ (*tf.gr*, *tf.ig*, *prob-based*).

On the la12 corpus, *icf-based* $>>$ (*tf.rf*, *tf.icf*, *tf.idf*) $>>$ *tf.logOR* $>>$ (*tf.* $\chi^2$, *tf.ig*, *tf.gr*) $>>$ *prob-based* when employing SVM or centroid-based classifier; The ranking order is *icf-based* $>>$ (*tf.rf*, *tf.icf*) $>>$ *tf.idf* $>>$ (*tf.* $\chi^2$, *tf.logOR*) $>>$ (*tf.ig*, *tf.gr*) $>>$ *prob-based* when kNN is used.

The results above indicate that our proposed *icf-based* method is superior or comparable to existing term weighting methods, especially when centroid classifier or kNN is used.

### 6.3 Discussions

Through experiments we can find that our novel *tf.icf* and *icf-based* STW have improved the performance of text categorization for multi-class, and binary classification, respectively, compared with other existing methods. The performance of these term weighting schemes can be summarized as follows:

- Inverse category frequency can reflect the distribution of terms among categories, and it is useful for TC task. Our novel *tf.icf* and *icf-based* supervised term weighting scheme outperform eight existing methods in the controlled experiments, such as *tf.rf*, *prob-based* and *tf.idf*.
- Especially for *tf.icf*, we suggest it as the standard term weighting scheme for **multi-class classification tasks**, because *icf* gives more reasonable explanation than *idf*, and *tf.icf* consistently outperforms *tf.idf*.
- For **binary classification tasks**, supervised term weighting schemes, which adopt the known categorical information, can obtain better performance. Our *icf-based*, combining *rf* and *icf*, can outperform existing supervised term weighting schemes on three widely used corpora in most cases. And the improvement is significant when similarity-based classifiers are used, such as kNN and centroid classifiers.
- However, traditional *tf.idf* also has its own superiority and outperforms some supervised term weighting schemes (e.g., *prob-based*, *tf.* $\chi^2$, and *tf.ig*).

We should point out that the conclusions above are made in combination with SVM, kNN, and centroid-based classifiers in terms of *macro-F1* and *micro-F1* and other controlled settings.

## 7. CONCLUSIONS

Compared with unsupervised term weighting methods, supervised ones for TC task have become an important research topic. The known categorical information of terms should be fully applied to text categorization. We introduce inverse category frequency into term weighting schemes, and propose *tf.icf* and *icf-based* supervised term weighting scheme, which combines inverse category frequency and relevance frequency. The introduction of *icf* can assign less weight to terms occurring in many categories. Our experimental results and extensive comparisons based on three common corpora, i.e., skewed Reuters-21578, balanced 20 Newsgroup and la12, have shown that our two term weighting schemes achieve the better or comparable performance than seven supervised term weighting schemes and three traditional term weighting schemes. For future work, we are going to conduct more experiments to validate the generalization of inverse category frequency for TC task.